# Measuring Plagiarism in Introductory Programming Course Assignments

Muhammad Humayoun
mhumayoun@hct.ac.ae

Muhammad Adnan Hashmi
ahashmi@hct.ac.ae

Ali Hanzala Khan
ahanzala@hct.ac.ae

Faculty of Computer Information Science
Abu Dhabi Men's Campus, Higher Colleges of Technology
United Arab Emirates

*Abstract*—Measuring plagiarism in programming assignments is an essential task to the educational procedure. This paper discusses the methods of plagiarism and its detection in introductory programming course assignments written in C++. A small corpus of assignments is made publicly available. A general framework to compute the similarity between a solution pair is developed that uses the three token-based similarity methods as features and predicts if the solution is plagiarized. The importance of each feature is also measured, which in return ranks the effectiveness of each method in use. Finally, the artificially generated dataset improves the results compared to the original data. We achieved an F1 score of 0.955 and 0.971 on original and synthetic datasets.

## I. Introduction

Copying someone else work without acknowledging the source is known as plagiarism. Plagiarism is a false claim of academic achievement that does not belong to a student. Therefore, plagiarism is considered a serious offense in academia, and it compromises academic integrity. In addition, programming jobs are high in demand, and the market need for good programmers is ever increasing. Students graduating with incompetent coding skills may undermine the success of tech companies that hire them. Coding is such an important skill that initiatives such as "One Million Arab Coders"[1] are launched to produce competent local programmers for the UAE's rising tech industry. Plagiarism in the introductory programming course assignments can start a habit formation of academic misconduct that can undermine all the needs and initiatives mentioned above. It is also unfair to those who produced and submitted original works. Moreover, it compromises the quality and reputation of a degree awarding institution.

This study tries to understand plagiarism in introductory programming courses using a data-driven approach. Introductory programming course assignments are unique with respect to the inherent higher similarity that two independent solutions may have. It is mainly because, first, the solutions are generally short; after all, these are easy problems. Second, there is a limited variation possible in solutions as students use limited syntax and constructs.

We present three key contributions. Firstly, we report the development of **I**ntroductory **P**rogramming **C**ourse **A**ssignments Corpus (IPCA). It is a small benchmark corpus containing 60 assignments, manually written by four volunteers (including one author). Among these assignments, there are 24 non-plagiarized and 36 plagiarized assignments. These assignments are taken from the first programming course in an undergraduate program taught in C++ in a university[2]. There are 10 versions per assignment, 4 of these are non-plagiarized, and 6 of these are plagiarized. We have publically released this corpus as open source[3]. Secondly, we produce a general framework to compute the similarity between programming assignments written in C++. Thirdly, we perform several benchmark experiments on the corpus, such as:

1. Computing similarity scores between plagiarized and non-plagiarized assignments using the token matching methods reported in literature.
2. Using these similarity scores as features for the supervised learning classifiers, we compute the likelihood of a target assignment being plagiarized or non-plagiarized.
3. The importance of each feature is measured using an attribute selection algorithm. This, in return, informs the effectiveness of the methods for introductory programming course assignments.
4. The dataset produced from IPCA is not balanced as we have 3.38 times more non-plagiarized instances than plagiarized ones. Therefore, a synthetic yet balanced dataset is produced from the IPCA corpus. In addition, its effectiveness for the task at hand is evaluated.

## II. Related Work

Systems such as JPLAG [1], MOSS [2], YAP [3], and SHERLOCK [4] are widely used for identifying plagiarism in students' provided solutions [5]. There have been many studies that evaluate the plagiarism detection systems, such as [6] [7] [8] [9] and [10], to mention a few. Plagiarism detection in source code can be categorized into four broad approaches. These approaches are *metrics based*, *token based*, *graph based*, and *abstract syntax tree based* techniques. In metrics based approach, different attributes of a source code such as number of lines, number of unique words [11], number of operators and operands, number of distinct operators, and operands [12] are considered. These attributes are compared between a pair of coding assignments using some similarity function.

In a token based approach, the code is converted into a standard set of tokens. The tokens of source codes are then matched to determine their similarity using a similarity function [13]. The Token based approach is adapted by the tools Yap3, Sherlock, and JPlag [13]. However, MOSS adapts a little different technique named Winnowing [14] in which fingerprints of a solution is generated. According to [15], token matching approach is adapted widely due to its simplicity, efficiency, and accuracy.

---

[1] One Million Arab Coders. Dubai Future Foundation - United Arab Emirates. https://www.arabcoders.ae/ Last accessed March 15, 2022.

[2] The University of Central Punjab. Lahore, Pakistan. https://ucp.edu.pk/
[3] GitHub link: https://github.com/humsha/IPCA

In Abstract Syntax Tree (AST) based approach, the source codes are first converted in their respective ASTs. Then the algorithm searches for sub-tree similarities between them. The work of [16] [17] [18] are some examples. In the graph based approach, the source code is first converted into a Program Dependency Graph (PDG). The subsets of PDGs of source files are then compared to find similarities [19].

Similar to our work, some studies use machine learning approach to decide whether two solutions are plagiarized or not. For instance, the work [20] utilizes a neural network technique to detect plagiarism and also rank various features for detecting plagiarism. In work [21], several machine learning methods are utilized with a number of feature pairs. The study [22] deals with plagiarism detection in assignments written in C. Machine learning and deep learning methods are investigated. Our work is unique because we focus on programming assignments written for the introductory course. We also convert the source solutions to ASTs, generalizing well, resulting in better similarity.

### A. Similarity functions

The most common similarity function in the token based approach is Greedy String Tiling [23]. In text plagiarism the most common similarity functions can be the Ngram overlap approach [24] [25] [26] [27] and Longest Common Subsequence (LCS) [25] [28]. In this study, we use these three similarity functions. Ngram and LCS are investigated because these are perhaps the simplest yet effective approaches to the problem. Since we convert the source solutions to ASTs (a better generalization), we think that these similarity functions can get better similarity scores.

#### 1) Ngram Overlap

A simple approach to compute similarity between two coding solutions is to measure the number of shared token ngrams. Each coding solution is treated as a set of overlapping n-token sequences and computes a similarity score from this. We particularly compute unigram, bigram and, trigram (ngram where n=1, 2, and, 3 respectively) scores between two coding solutions.

#### 2) Longest Common Subsequence

A longest common subsequence (LCS) is a common subsequence of maximal length. The similarity between a pair of solutions is calculated as the number of LCS in a pair divided by the total tokens in the source solution.

#### 3) Greedy String Tiling

Greedy String Tiling (GST) computes the similarity between two sequences and have been used in software code, free text or biological subsequences [23]. In contrast to other methods, GST can deal with the transposition of tokens [29].

## III. BENCHMARK CORPUS

We have collected six programming assignments generally given to students in the introductory programming course. The assignments are written in C++. These assignments are the following:

1. Swapping the first and the last element of an array
2. Computing Factorial of a number taken as input
3. Checking if a number taken as input is prime or not
4. Reverse a number taken as input
5. Addition of two numbers using function taken as input
6. Printing even or odd numbers from 1 to 50 based on a switchboard

For each assignment, there are four non-plagiarized solutions written by four volunteers. Then, the last two versions are plagiarized. The volunteers are asked to plagiarize these assignment copies as follows:

1. Plagiarizing a copy by adding comments
2. Plagiarizing a copy by change of variable names
3. Plagiarizing a copy by change of loops type, for instance, *for* loop to *while* loop and vice versa.

According to [30], there can be six levels of plagiarism found in programing assignments[4]. We have picked level one, three, and five for our plagiarized versions as these levels seem to be more appropriate for the first programming course.

The volunteers were the students of the computer science undergraduate program studying in the last semester. The corpus generation task was conducted in one session (2 hours work) with three volunteer students and one author. Sample 3 from Assignment 1 and its two plagiarized versions are shown in Figure 1.

## IV. FRAMEWORK FOR PLAGIARISM DETECTION

In order to recognize valid C++ tokens such as numbers, strings, arrays, functions, operators, proper tokenization needs to be applied. However, this tokenization cannot be simply performed on whitespaces. A lexer program is needed that recognizes the syntax and keywords associated with the programming language.

The BNFC[5] convertor tool [31] takes a Labelled BNF (Backus Normal Form) grammar[6] [32] as input and generates a compiler front, which is capable of producing abstract syntax trees. Abstract syntax trees abstract away from whitespaces, additional lines, and coding comments. Compiler frontends generated by BNFC tool have been used not only for programming languages but also for other formalisms such as the language of mathematics and the first order logic [33] [34] [35] [36].

We develop a Labelled BNF grammar for C++ and used the BNFC convertor tool to generate a compiler front end. As a first step, the solution of a coding assignment is converted into its abstract syntax tree. The abstract syntax tree and the linearized tree of code segment containing *for* loop as an example are shown in Figure 1. Since the coding comments are automatically removed from the abstract syntax tree, a plagiarized copy (produced by adding comments only) has a

---

[4] Level 0: An Exact copy of original source code; Level 1: Modification of comments; Level 2: Modification of variable names; Level 3: Changing scope of variables; Level 4: Transformation of several segments into procedures; Level 5: Change of the structure i.e. from for loop to while; Level 6: Change the logic.

[5] https://bnfc.digitalgrammars.com/
[6] http://bnfc.digitalgrammars.com/LBNF-report.pdf

Figure 1. The Sample 3 from Assignment 1 and its two plagiarized versions (Swapping the first and the last element of an array)

| Solution3.cpp<br>**Non Plagiarized** | Solution3.cpp<br>**Plagiarized change in Loop type** | Solution3.cpp<br>**Plagiarized change in variable names** |
|---|---|---|
| ```cpp
#include<iostream.h>
#include<conio.h>
int main(){
   int Arr[100],n,temp;
   cout<<"Enter # of elements you want to insert ";
   cin>>n;
   for(int i=0;i<n;i=i+1)
   {
    cout<<"Enter element "<<i+1<<":";
    cin>>Arr[i];
   }
   temp=Arr[0];
   Arr[0]=Arr[n-1];
   Arr[n-1]=temp;
   cout<<"\nArray after swapping"<<endl;
   for(i=0;i<n;i=i+1)
      cout<<Arr[i]<<" ";
   return 0;
}
``` | ```cpp
#include<iostream.h>
#include<conio.h>
int main(){
   int Arr[100],n,temp;
   cout<<"Enter # of elements you want to insert";
   cin>>n;
   int i=0;
   while(i<n)
   {
    cout<<"Enter element"<<i+1<<":";
    cin>>Arr[i];
    i=i+1;
   }
   temp=Arr[0];
   Arr[0]=Arr[n-1];
   Arr[n-1]=temp;
   cout<<"\nArray after swapping"<<endl;
   int j=0;
   while(j<n)  {
       cout<<Arr[j]<<" ";
       j=j+1;
   }
  return 0;
}
``` | ```cpp
#include<iostream.h>
#include<conio.h>
int main(){
    int array1[100],num,temp;
    cout<<"Enter # of elements you want to insert ";
    cin>>num;
    for(int i=0;i<num;i=i+1)
    {
     cout<<"Enter element "<<i+1<<":";
     cin>>array1[i];
    }
    temp=array1[0];
    array1[0]=array1[num-1];
    array1[num-1]=temp;
    cout<<"\narray1 after swapping"<<endl;
    for(i=0;i<num;i=i+1)
       cout<<array1[i]<<" ";
    return 0;
}
``` |

**Code segment containing for loop from Solution3.cpp:**
```
for(i=0;i<n;i=i+1)
     cout << Arr [i]<< " ";
```

**[Abstract Syntax Tree]**
```
(IterS [(Sfor (Sexpr [(Eassign (Evar "i") [Assign] (Econst [(Eint 0)])))]) (Sexpr [(Elessthen (Evar "i") (Evar "n"))] ) [(Eassign (Evar "i") [Assign] (Eplus (Evar "i") (Econst [(Eint 1)])))] [(ExprS [(Sexpr [(Eleft (Eleft (Oper "cout") (Earray (Evar "Arr") (Evar "i") )) (Estring " "))] )])])
```

**[Linearized Tree]**
```
for (i  =  0;
    i  <  n;
    i  =  i  + 1)
       cout << Arr [i]<< " ";
```

---

similarity score of 100%. Second, similarity scores of two assignment pairs (on the corresponding abstract syntax trees) are computed. The algorithms we use are Ngram overlap approach (with n=1, 2, 3), Longest Common Subsequence, and Greedy String Tiling (with n=2, 3).

*A. Similarity scores for assignments*

*1) Similarity in Non-Plagiarized assignments*

Table I discusses the similarity results in non-plagiarized assignments. It is worth noting that, overall, the similarity scores for introductory programming course assignments are generally higher. The highest scores (i.e., 88.4) are found in assignment 6, whereas the lowest scores (i.e., 71.8) are in assignment 5.

*2) Similarity in All assignments*

Table II showcases the similarity results in all assignments. It is worth noting that comments from a solution are completely removed when parsed by the lexer. Therefore, a plagiarized copy produced by adding comments only, have a similarity score of 100% with the corresponding non-plagiarized solution. Because of that, we omit all the instances of plagiarized copies made by adding comments. Another point worth noting is that the plagiarized solutions by change of variables or loop have the highest similarity scores with the corresponding non-plagiarized solutions. Results for assignment 1 in Table II have the minimum similarity score of 55.9.

For assignment 2, the minimum similarity score is 55.2 between two non-plagiarized solutions written independently. For the remaining assignments, we only show the top results. It can be observed that the similarity scores between different similarity measures tend to vary from each other. For instance, consider the similarity score between the solutions *A1-NP-Sol-1* and *A1-P-Sol-4-variables* in Table II (SN:1, the highest

TABLE I. Similarity in Non-Plagiarized assignments. NP: non-plagiarized; P: plagiarized; LCS: Longest Common Subsequence; N1, N2, N3: N-gram overlap n=1,2,3; GST1, GST2, GST3: Greedy String Tiling n=1,2,3 respectively. Results are sorted on AVG from highest to lowest.

Assignment 1: Swapping the first and the last element of an array.

| SN | Solution i | Solution j | LCS | N1 | N2 | N3 | GST1 | GST2 | GST3 | AVG | STDV |
|---|---|---|---|---|---|---|---|---|---|---|---|
| 1 | A1-NP-Sol-1 | A1-NP-Sol-4 | 79 | 81.5 | 76 | 65.8 | 80.5 | 79.6 | 78.7 | **77.3** | 5.4 |
| 2 | A1-NP-Sol-1 | A1-NP-Sol-3 | 67.5 | 68.7 | 60.2 | 47.9 | 90.5 | 87.9 | 85.6 | **72.6** | 16 |
| 3 | A1-NP-Sol-3 | A1-NP-Sol-4 | 68.6 | 69.8 | 56.8 | 45.5 | 89.3 | 85.3 | 82.8 | **71.2** | 16 |
| 4 | A1-NP-Sol-2 | A1-NP-Sol-3 | 61.9 | 66.3 | 55.5 | 45.7 | 87 | 84.3 | 82.2 | **69** | 15.9 |
| 5 | A1-NP-Sol-2 | A1-NP-Sol-4 | 57.3 | 65.2 | 51.7 | 38.7 | 94.1 | 88.4 | 84.8 | **68.6** | 20.9 |
| 6 | A1-NP-Sol-1 | A1-NP-Sol-2 | 66.3 | 68.6 | 56.1 | 41.2 | 78.3 | 77.9 | 77.4 | **66.5** | 13.8 |

Assignment 2: Computing Factorial of a number taken as input

| SN | Solution i | Solution j | LCS | N1 | N2 | N3 | GST1 | GST2 | GST3 | AVG | STDV |
|---|---|---|---|---|---|---|---|---|---|---|---|
| 1 | A2-NP-Sol-3 | A2-NP-Sol-4 | 72.6 | 72.6 | 61.9 | 51.2 | 92 | 88.8 | 87.8 | **75.3** | 15.2 |
| 2 | A2-NP-Sol-2 | A2-NP-Sol-4 | 66.7 | 69.5 | 58 | 45.3 | 96.8 | 91.4 | 90 | **74** | 19.3 |
| 3 | A2-NP-Sol-2 | A2-NP-Sol-3 | 67.5 | 71.4 | 61.8 | 53.1 | 90 | 87 | 84.6 | **73.6** | 14 |
| 4 | A2-NP-Sol-1 | A2-NP-Sol-3 | 43.1 | 55.1 | 41.5 | 29.7 | 89.1 | 79.2 | 73.5 | **58.7** | 22.2 |
| 5 | A2-NP-Sol-1 | A2-NP-Sol-2 | 49 | 54.3 | 41.9 | 31.1 | 81.4 | 76.6 | 74.3 | **58.4** | 19.3 |
| 6 | A2-NP-Sol-1 | A2-NP-Sol-4 | 42.9 | 48.1 | 37.1 | 27.3 | 83.1 | 75.9 | 72.1 | **55.2** | 21.6 |

Assignment 3: Checking if a number taken as input is prime or not

| SN | Solution i | Solution j | LCS | N1 | N2 | N3 | GST1 | GST2 | GST3 | AVG | STDV |
|---|---|---|---|---|---|---|---|---|---|---|---|
| 1 | A3-NP-Sol-1 | A3-NP-Sol-4 | 80.3 | 81.6 | 66.1 | 55.4 | 96.2 | 94.5 | 93.8 | **81.1** | 15.6 |
| 2 | A3-NP-Sol-3 | A3-NP-Sol-4 | 66.2 | 74.5 | 59.5 | 45.2 | 95.8 | 90.3 | 87.7 | **74.2** | 18.4 |
| 3 | A3-NP-Sol-2 | A3-NP-Sol-3 | 66.3 | 73.6 | 57.1 | 44.4 | 88.9 | 85.7 | 83 | **71.3** | 16.4 |
| 4 | A3-NP-Sol-1 | A3-NP-Sol-3 | 68.5 | 71.1 | 51.1 | 34.5 | 95.1 | 88.6 | 85.1 | **70.6** | 21.7 |
| 5 | A3-NP-Sol-2 | A3-NP-Sol-4 | 66.3 | 68.7 | 52.5 | 42.5 | 87.4 | 84.9 | 83.3 | **69.4** | 17.2 |
| 6 | A3-NP-Sol-1 | A3-NP-Sol-2 | 62.4 | 67.1 | 45.2 | 33.6 | 89.5 | 85.1 | 82.9 | **66.5** | 21.2 |

Assignment 4: Reverse a number taken as input

| SN | Solution i | Solution j | LCS | N1 | N2 | N3 | GST1 | GST2 | GST3 | AVG | STDV |
|---|---|---|---|---|---|---|---|---|---|---|---|
| 1 | A4-NP-Sol-1 | A4-NP-Sol-3 | 78.3 | 78.3 | 73.1 | 64.6 | 97.2 | 95.5 | 94.6 | **83.1** | 12.7 |
| 2 | A4-NP-Sol-1 | A4-NP-Sol-4 | 68.4 | 70.1 | 56.1 | 41.1 | 72.6 | 71.5 | 71 | **64.4** | 11.7 |
| 3 | A4-NP-Sol-1 | A4-NP-Sol-2 | 45.4 | 66.3 | 53.1 | 42.8 | 82 | 80 | 76.8 | **63.8** | 16.6 |
| 4 | A4-P-Sol-3 | A4-NP-Sol-4 | 61.7 | 63.3 | 51.2 | 35.8 | 71.2 | 69.9 | 68.9 | **60.3** | 12.8 |
| 5 | A4-NP-Sol-2 | A4-NP-Sol-3 | 42.2 | 55.4 | 41 | 29.9 | 83.2 | 79.1 | 75 | **58** | 21.2 |
| 6 | A4-NP-Sol-2 | A4-NP-Sol-4 | 40 | 50 | 36 | 24.8 | 56.8 | 55.9 | 55.5 | **45.6** | 12.3 |

Assignment 5: Addition of two numbers using function taken as input

| SN | Solution i | Solution j | LCS | N1 | N2 | N3 | GST1 | GST2 | GST3 | AVG | STDV |
|---|---|---|---|---|---|---|---|---|---|---|---|
| 1 | A1-NP-Sol-1 | A1-NP-Sol-2 | 64.2 | 67.9 | 53.9 | 46.4 | 92.7 | 89.4 | 87.8 | **71.8** | 18.4 |
| 2 | A1-NP-Sol-1 | A1-NP-Sol-3 | 52.3 | 65.8 | 56.7 | 46.3 | 92.6 | 84.4 | 82.1 | **68.6** | 17.9 |
| 3 | A1-NP-Sol-2 | A1-NP-Sol-3 | 45.5 | 59.7 | 48 | 39.1 | 89.8 | 83.2 | 81.6 | **63.8** | 20.7 |
| 4 | A1-NP-Sol-1 | A1-NP-Sol-4 | 48 | 53.3 | 39.6 | 32 | 92.6 | 86.8 | 83.9 | **62.3** | 24.8 |
| 5 | A1-NP-Sol-2 | A1-NP-Sol-4 | 43.9 | 51.6 | 39.6 | 32.5 | 95 | 88.4 | 85.3 | **62.3** | 26.3 |
| 6 | A1-NP-Sol-3 | A1-NP-Sol-4 | 48.3 | 59.3 | 43.4 | 34.4 | 88.3 | 81 | 76.8 | **61.6** | 20.7 |

Assignment 6: Printing even or odd numbers from 1 to 50 based on a switch board

| SN | Solution i | Solution j | LCS | N1 | N2 | N3 | GST1 | GST2 | GST3 | AVG | STDV |
|---|---|---|---|---|---|---|---|---|---|---|---|
| 1 | A2-NP-Sol-3 | A2-NP-Sol-4 | 89.7 | 89.7 | 83.4 | 75.1 | 95.1 | 93.4 | 92.4 | **88.4** | 7 |
| 2 | A2-NP-Sol-2 | A2-NP-Sol-3 | 88.9 | 88.9 | 80.2 | 75.9 | 96.2 | 93.1 | 91.5 | **87.8** | 7.2 |
| 3 | A2-NP-Sol-1 | A2-NP-Sol-3 | 81.6 | 88.5 | 77.1 | 67.2 | 94 | 90.6 | 87.5 | **83.8** | 9.2 |
| 4 | A2-NP-Sol-2 | A2-NP-Sol-4 | 84.5 | 84.5 | 74.3 | 66.9 | 93.7 | 91 | 89.8 | **83.5** | 9.7 |
| 5 | A2-NP-Sol-1 | A2-NP-Sol-4 | 75.1 | 80.9 | 66.7 | 56.7 | 91.7 | 88 | 84.6 | **77.7** | 12.4 |
| 6 | A2-NP-Sol-1 | A2-NP-Sol-2 | 67.1 | 78.8 | 62.3 | 53.1 | 94.9 | 89.4 | 85.3 | **75.8** | 15.4 |

score). The second solution is not a plagiarized version of the first solution, but the similarity is the highest. The same scenario is for the serial numbers 7, 8, 9, 10, 12. The similarity scores in assignments 2 to 6 in Table II, have the same tendencies.

## V. EXPERIMENTS

In Section IV-2, it is observed that the similarity scores between plagiarized solutions and non-plagiarized solutions are kind of mixed. A careful analysis is needed to find the ideal score separating plagiarized and non-plagiarized copies. Therefore, to evaluate the approaches for measuring similarity in solutions, we cast the problem into a supervised learning task. We used similarity scores as attributes and 'plagiarized' and 'non-plagiarized' as labels (the correct answers) for machine learning algorithms (Classifiers). Once a classifier is

TABLE II. Similarity in all assignments. NP: non-plagiarized; P: plagiarized; LCS: Longest Common Subsequence; N1, N2, N3: N-gram overlap n=1,2, 3; GST1, GST2, GST3: Greedy String Tiling n=1, 2, 3 respectively. Results are sorted on AVG from highest to lowest.

Assignment 1: Swapping the first and the last element of an array

| SN | Solution i | Solution j | LCS | N1 | N2 | N3 | GST1 | GST2 | GST3 | AVG | STDV |
|---|---|---|---|---|---|---|---|---|---|---|---|
| 1 | A1-NP-Sol-1 | A1-P-Sol-4-variables | 78 | 80.5 | 75.8 | 68.5 | 80.8 | 79.7 | 78.8 | **77.4** | 4.3 |
| 2 | A1-NP-Sol-3 | A1-P-Sol-3-variables | 93 | 93 | 89.3 | 84.8 | 99 | 99 | 99 | **93.9** | 5.5 |
| 3 | A1-NP-Sol-3 | A1-P-Sol-3-loops | 91.4 | 92.6 | 86.5 | 80 | 95 | 93.1 | 91.9 | **90.1** | 5.2 |
| 4 | A1-NP-Sol-4 | A1-P-Sol-4-variables | 88.1 | 88.1 | 80.6 | 69.8 | 98.7 | 98.2 | 98.2 | **88.8** | 10.8 |
| 5 | A1-P-Sol-3-loops | A1-P-Sol-3-variables | 83.4 | 85.7 | 76.6 | 66.9 | 94.5 | 92.1 | 90.6 | **84.3** | 9.7 |
| 6 | A1-NP-Sol-4 | A1-P-Sol-4-loops | 92.5 | 92.5 | 85.2 | 79.5 | 72.7 | 72.3 | 72.1 | **81** | 9.2 |
| 7 | A1-NP-Sol-1 | A1-P-Sol-3-variables | 67.5 | 68.7 | 60.2 | 48.1 | 89.8 | 87.8 | 85.6 | **72.5** | 15.8 |
| 8 | A1-NP-Sol-3 | A1-P-Sol-4-variables | 69 | 70.2 | 58.5 | 49.4 | 88.8 | 85.4 | 82.7 | **72** | 14.6 |
| 9 | A1-P-Sol-3-variables | A1-P-Sol-4-variables | 67.8 | 69 | 57.8 | 49.6 | 89.2 | 85.6 | 82.2 | **71.6** | 14.8 |
| 10 | A1-P-Sol-4-loops | A1-P-Sol-4-variables | 81.1 | 81.1 | 67.1 | 53.6 | 72.8 | 71 | 71 | **71.1** | 9.4 |
| 11 | A1-P-Sol-3-variables | A1-NP-Sol-4 | 67.5 | 68.6 | 56.1 | 45.6 | 89.5 | 86 | 83.1 | **70.9** | 16.3 |
| 12 | A1-P-Sol-3-loops | A1-P-Sol-4-variables | 62.1 | 67.8 | 53.8 | 44.8 | 92.3 | 85.7 | 81.2 | **69.7** | 17.5 |
| 13 | A1-NP-Sol-2 | A1-P-Sol-3-variables | 61.9 | 66.3 | 55.5 | 45.8 | 87.8 | 85.1 | 83.1 | **69.4** | 16.3 |
| 14 | A1-NP-Sol-2 | A1-P-Sol-4-variables | 57.8 | 65.6 | 52.7 | 41.5 | 93.9 | 87.1 | 83.2 | **68.8** | 19.6 |
| 15 | A1-NP-Sol-1 | A1-P-Sol-3-loops | 58.9 | 66.3 | 54.4 | 43.5 | 86.9 | 83.4 | 81.6 | **67.9** | 16.6 |
| 16 | A1-P-Sol-3-loops | A1-NP-Sol-4 | 59.3 | 66.3 | 49.1 | 37.2 | 92.4 | 85.6 | 81.1 | **67.3** | 20.2 |
| 17 | A1-P-Sol-Sol-2 | A1-P-Sol-3-loops | 53.3 | 64.1 | 49.8 | 37.1 | 91 | 87.5 | 85.4 | **66.9** | 21.3 |
| 18 | A1-NP-Sol-1 | A1-NP-Sol-2 | 66.3 | 68.6 | 56.1 | 41.2 | 78.3 | 77.9 | 77.4 | **66.5** | 13.8 |
| 19 | A1-P-Sol-3-loops | A1-P-Sol-4-loops | 65.9 | 69.3 | 56.2 | 45.8 | 66.7 | 65.7 | 63.9 | **61.9** | 8.2 |
| 20 | A1-NP-Sol-1 | A1-P-Sol-4-loops | 72 | 74.4 | 63.4 | 51.3 | 55.9 | 55.2 | 54.4 | **60.9** | 9.2 |
| 21 | A1-NP-Sol-2 | A1-P-Sol-4-loops | 53 | 65.9 | 48.5 | 33.8 | 74.3 | 72.3 | 69.1 | **59.6** | 14.9 |
| 22 | A1-NP-Sol-3 | A1-P-Sol-4-loops | 64.8 | 65.9 | 51 | 39.7 | 63.1 | 61.7 | 60 | **58** | 9.5 |
| 23 | A1-P-Sol-3-variables | A1-P-Sol-4-loops | 60.2 | 62.5 | 46.9 | 35.9 | 63.8 | 62.3 | 60 | **55.9** | 10.5 |

Assignment 2: Computing Factorial of a number taken as input

| SN | Solution i | Solution j | LCS | N1 | N2 | N3 | GST1 | GST2 | GST3 | AVG | STDV |
|---|---|---|---|---|---|---|---|---|---|---|---|
| 1 | A2-P-Sol-4 | A2-P-Sol-4-variables | 94.4 | 94.4 | 90.3 | 83.5 | 99.2 | 98.9 | 98.9 | 94.2 | 5.8 |
| 2 | A2-P-Sol-3 | A2-P-Sol-3-loops | 92.9 | 95.2 | 89.3 | 84.8 | 97.1 | 95.3 | 94.7 | 92.8 | 4.3 |
| 3 | A2-NP-Sol-4 | A2-P-Sol-4-loops | 89.5 | 93.7 | 87.8 | 82.3 | 95.9 | 94.1 | 93.3 | 90.9 | 4.7 |
| 4 | A2-P-Sol-4-loops | A2-P-Sol-4-variables | 82.5 | 88.1 | 79 | 68.4 | 95.4 | 92.9 | 91.7 | 85.4 | 9.5 |
| 5 | A2-NP-Sol-3 | A2-P-Sol-3-variables | 79.8 | 79.8 | 71.9 | 61.9 | 96.7 | 94.9 | 94.5 | 82.8 | 13.2 |
| 6 | A2-NP-Sol-3 | A2-P-Sol-4-variables | 77.7 | 77.7 | 70.8 | 63.2 | 91.6 | 89.7 | 88.7 | 79.9 | 10.7 |
| 7 | A2-P-Sol-3-loops | A2-P-Sol-3-variables | 72.3 | 75.9 | 63.9 | 52.4 | 96.9 | 92.7 | 91.4 | 77.9 | 16.6 |
| 8 | A2-NP-Sol-2 | A2-P-Sol-4-variables | 70.9 | 73.8 | 64 | 53.3 | 97.3 | 92.4 | 91.2 | 77.6 | 16.5 |
| 9 | A2-P-Sol-3-variables | A2-NP-Sol-4 | 71 | 71 | 63.1 | 54.8 | 89.8 | 88 | 87.1 | 75 | 13.6 |
| 10 | A2-NP-Sol-2 | A2-NP-Sol-4 | 66.7 | 69.5 | 58 | 45.3 | 96.8 | 91.4 | 90 | 74 | 19.3 |
| 11 | A2-NP-Sol-2 | A2-NP-Sol-3 | 67.5 | 71.4 | 61.8 | 53.1 | 90 | 87 | 84.6 | 73.6 | 14 |
| 12 | A2-P-Sol-3-loops | A2-P-Sol-4-loops | 70.1 | 70.1 | 57.9 | 47.2 | 92.7 | 88.8 | 87.7 | 73.5 | 17.1 |
| 13 | A2-P-Sol-3-loops | A2-P-Sol-4-variables | 65.8 | 72.3 | 62.2 | 53 | 89.7 | 86.8 | 84.6 | 73.5 | 14 |
| 14 | A2-P-Sol-3-variables | A2-P-Sol-4-variables | 68.4 | 68.4 | 59.5 | 50.8 | 89 | 87.6 | 86.6 | 72.9 | 15.1 |
| 15 | A2-NP-Sol-2 | A2-P-Sol-3-loops | 63.2 | 69.7 | 59.3 | 52.5 | 88.1 | 86.2 | 84.5 | 71.9 | 14.4 |
| 16 | A2-NP-Sol-2 | A2-P-Sol-4-loops | 62.9 | 67.1 | 55.3 | 44.2 | 94.2 | 90.2 | 88.3 | 71.7 | 19.4 |
| 17 | A2-P-Sol-3-variables | A2-P-Sol-4-loops | 62.3 | 66.2 | 56.1 | 47.4 | 92.4 | 88.2 | 85.4 | 71.1 | 17.5 |
| 18 | A2-P-Sol-3 | A2-P-Sol-4-loops | 64.1 | 66.7 | 55.3 | 41.4 | 94 | 87.7 | 84.8 | 70.3 | 19.4 |
| 19 | A2-P-Sol-3-loops | A2-NP-Sol-4 | 61.9 | 67.1 | 54.1 | 42.6 | 90 | 85.3 | 83.8 | 69.3 | 17.8 |
| 20 | A2-NP-Sol-2 | A2-P-Sol-3-variables | 61.8 | 64.5 | 51.9 | 40.3 | 87.6 | 84.7 | 82.2 | 67.6 | 18 |
| 21 | A2-NP-Sol-1 | A2-P-Sol-3-loops | 43.6 | 54.5 | 40.5 | 29.9 | 91 | 82.7 | 78.1 | 60 | 23.8 |
| 22 | A2-NP-Sol-1 | A2-NP-Sol-3 | 43.1 | 55.1 | 41.5 | 29.7 | 89.1 | 79.2 | 73.5 | 58.7 | 22.2 |
| 23 | A2-NP-Sol-1 | A2-NP-Sol-2 | 49 | 54.3 | 41.9 | 31.1 | 81.4 | 76.6 | 74.3 | 58.4 | 19.3 |
| 24 | A2-NP-Sol-1 | A2-P-Sol-3-variables | 41.2 | 50.9 | 37.1 | 26.2 | 90.6 | 80.2 | 74.6 | 57.3 | 24.5 |
| 25 | A2-NP-Sol-1 | A2-P-Sol-4-loops | 41.8 | 47.1 | 36.4 | 27.2 | 86.1 | 79.4 | 75.1 | 56.2 | 23.5 |
| 26 | A2-NP-Sol-1 | A2-P-Sol-4-variables | 42.9 | 49.4 | 38.9 | 28.9 | 82.3 | 76.4 | 72.4 | 55.9 | 20.9 |
| 27 | A2-NP-Sol-1 | A2-NP-Sol-4 | 42.9 | 48.1 | 37.1 | 27.3 | 83.1 | 75.9 | 72.1 | 55.2 | 21.6 |

Assignment 3: Checking if a number taken as input is prime or not

| SN | Solution i | Solution j | LCS | N1 | N2 | N3 | GST1 | GST2 | GST3 | AVG | STDV |
|---|---|---|---|---|---|---|---|---|---|---|---|
| 1 | A3-NP-Sol-3 | A3-P-Sol-3-loops | 94.5 | 94.5 | 92.3 | 90.6 | 99.5 | 99.1 | 99.1 | 95.7 | 3.6 |
| 2 | A3-NP-Sol-4 | A3-P-Sol-4-loops | 92.5 | 92.5 | 85.8 | 81.3 | 96.4 | 94.9 | 94.2 | 91.1 | 5.5 |
| 3 | A3-NP-Sol-4 | A3-P-Sol-4-variables | 90.5 | 90.5 | 82.9 | 71.9 | 98.6 | 98.6 | 98.6 | 90.2 | 10 |
| 4 | A3-P-Sol-3-loops | A3-P-Sol-3-variables | 90.7 | 90.7 | 80.5 | 69 | 97.4 | 96.1 | 96.1 | 88.6 | 10.4 |
| 5 | A3-NP-Sol-1 | A3-P-Sol-4-variables | 84.2 | 85.5 | 73.4 | 65.1 | 96.4 | 95.3 | 94.6 | 84.9 | 12 |
| 6 | A3-NP-Sol-3 | A3-P-Sol-3-variables | 84.9 | 84.9 | 72.7 | 60.3 | 96.9 | 95.4 | 95.1 | 84.3 | 13.6 |
| 7 | A3-P-Sol-4-loops | A3-P-Sol-4-variables | 81.6 | 83 | 69.8 | 56.9 | 96 | 93.8 | 92.6 | 82 | 14.3 |
| 8 | A3-NP-Sol-3 | A3-P-Sol-4-loops | 81.9 | 81.9 | 71 | 59.5 | 96.3 | 91.4 | 89.6 | 81.7 | 12.8 |

Assignment 4: Reverse a number taken as input

| SN | Solution i | Solution j | LCS | N1 | N2 | N3 | GST1 | GST2 | GST3 | AVG | STDV |
|---|---|---|---|---|---|---|---|---|---|---|---|
| 1 | A4-NP-Sol-4 | A4-P-Sol-4-loops | 94.8 | 94.8 | 92.7 | 90.1 | 98.1 | 98.1 | 98.1 | 95.2 | 3.1 |
| 2 | A4-NP-Sol-3 | A4-P-Sol-3-loops | 89.9 | 89.9 | 89 | 87.8 | 97.5 | 97.5 | 97.3 | 92.7 | 4.5 |
| 3 | A4-NP-Sol-4 | A4-P-Sol-4-variables | 93.6 | 93.6 | 88.3 | 76.6 | 98 | 98 | 98 | 92.3 | 7.8 |
| 4 | A4-P-Sol-4-loops | A4-P-Sol-4-variables | 88.7 | 88.7 | 81.3 | 67.2 | 97.3 | 96.4 | 96.1 | 88 | 10.8 |
| 5 | A4-NP-Sol-3 | A4-P-Sol-3-variables | 88.4 | 88.4 | 81.2 | 68.7 | 95.6 | 94.7 | 94.3 | 87.3 | 9.6 |
| 6 | A4-NP-Sol-1 | A4-P-Sol-3-loops | 83 | 83 | 78.7 | 71.9 | 96.8 | 95.8 | 95.4 | 86.4 | 9.7 |



**TABLE II.** Continued from the previous page.

Assignment 5: Addition of two numbers using function taken as input

| SN | Solution i | Solution j | LCS | N1 | N2 | N3 | GST1 | GST2 | GST3 | AVG | STDV |
|---|---|---|---|---|---|---|---|---|---|---|---|
| 1 | A1-P-Sol-4 | A1-P-Sol-4-variables | 84.9 | 84.9 | 77.6 | 71 | 99 | 99 | 99 | 87.9 | 11.4 |
| 2 | A1-P-Sol-3 | A1-P-Sol-3-variables | 79.2 | 79.2 | 71.4 | 64.2 | 97.5 | 94.8 | 94.8 | 83 | 12.9 |
| 3 | A1-NP-Sol-2 | A1-P-Sol-3-variables | 55.8 | 71.4 | 64 | 56.7 | 88.8 | 87.1 | 85.8 | 72.8 | 14.5 |
| 4 | A1-NP-Sol-1 | A1-P-Sol-4-variables | 57.3 | 62.7 | 52.5 | 46.4 | 93.2 | 87.6 | 85 | 69.2 | 18.9 |
| 5 | A1-NP-Sol-1 | A1-P-Sol-3-variables | 52.3 | 65.8 | 56.7 | 46.3 | 91.6 | 85.2 | 82.8 | 68.7 | 17.9 |
| 6 | A1-NP-Sol-2 | A1-P-Sol-4-variables | 43.9 | 51.6 | 39.6 | 32.5 | 95.1 | 88.4 | 85.3 | 62.3 | 26.3 |
| 7 | A1-P-Sol-3 | A1-P-Sol-4-variables | 48.3 | 59.3 | 43.4 | 34.4 | 88.5 | 80.8 | 76.9 | 61.7 | 20.7 |
| 8 | A1-P-Sol-3-variables | A1-P-Sol-4 | 48.3 | 59.3 | 43.4 | 34.4 | 87.3 | 80.8 | 77.5 | 61.6 | 20.6 |
| 9 | A1-P-Sol-3-variables | A1-P-Sol-4-variables | 48.3 | 59.3 | 43.4 | 34.4 | 87.5 | 80.8 | 77.5 | 61.6 | 20.6 |

Assignment 6: Printing even or odd numbers from 1 to 50 based on a switch board

| SN | Solution i | Solution j | LCS | N1 | N2 | N3 | GST1 | GST2 | GST3 | AVG | STDV |
|---|---|---|---|---|---|---|---|---|---|---|---|
| 1 | A2-P-Sol-3 | A2-P-Sol-3-variables | 94 | 94 | 90.1 | 83.2 | 97.6 | 96.9 | 96.9 | 93.2 | 5.1 |
| 2 | A2-P-Sol-4 | A2-P-Sol-4-variables | 93.9 | 93.9 | 90.2 | 83.6 | 96.5 | 95.9 | 95.9 | 92.8 | 4.6 |
| 3 | A2-P-Sol-3-variables | A2-P-Sol-4 | 92.1 | 92.1 | 87.9 | 83.8 | 96.4 | 95.9 | 95.4 | 91.9 | 4.7 |
| 4 | A2-P-Sol-3-loops | A2-P-Sol-4-loops | 89.4 | 89.4 | 81.9 | 75.6 | 97.1 | 94.4 | 93.6 | 88.8 | 7.6 |
| 5 | A2-P-Sol-3 | A2-P-Sol-4-variables | 89.7 | 89.7 | 83.4 | 75.1 | 94.9 | 94.4 | 93.9 | 88.7 | 7.2 |
| 6 | A2-P-Sol-3-variables | A2-P-Sol-4-variables | 89.7 | 89.7 | 83.4 | 75.1 | 93.1 | 92.1 | 91.6 | 87.8 | 6.4 |
| 7 | A2-P-Sol-4 | A2-P-Sol-4-loops | 85.2 | 89.9 | 81.7 | 74 | 96.3 | 93.5 | 91.5 | 87.4 | 7.7 |
| 8 | A2-NP-Sol-2 | A2-P-Sol-4-variables | 84.5 | 84.5 | 74.3 | 66.9 | 92.7 | 91.2 | 90.1 | 83.5 | 9.6 |
| 9 | A2-P-Sol-3-loops | A2-P-Sol-3-variables | 81.1 | 85.7 | 72.7 | 63.1 | 94.9 | 91.9 | 90.3 | 82.8 | 11.5 |
| 10 | A2-P-Sol-3 | A2-P-Sol-3-loops | 82.3 | 86.9 | 72.7 | 60.2 | 95 | 91.5 | 89.4 | 82.6 | 12.2 |
| 11 | A2-NP-Sol-2 | A2-P-Sol-3-variables | 82.7 | 82.7 | 71 | 62.9 | 95.1 | 91 | 89.2 | 82.1 | 11.5 |
| 12 | A2-P-Sol-4-loops | A2-P-Sol-4-variables | 79.3 | 84 | 74 | 63.7 | 94 | 90.5 | 87.9 | 81.9 | 10.5 |
| 13 | A2-NP-Sol-1 | A2-P-Sol-3-variables | 79.3 | 85.1 | 72 | 61.1 | 94.5 | 90.9 | 87.3 | 81.5 | 11.7 |
| 14 | A2-P-Sol-3-variables | A2-P-Sol-4-loops | 77.6 | 82.4 | 70.9 | 60 | 95.2 | 90.8 | 88.1 | 80.7 | 12.3 |
| 15 | A2-P-Sol-3-loops | A2-P-Sol-4 | 77 | 81.6 | 66.7 | 56.7 | 96.6 | 92.4 | 89.3 | 80 | 14.4 |
| 16 | A2-P-Sol-3 | A2-P-Sol-4-loops | 75.3 | 80 | 67.5 | 55.6 | 94.9 | 89 | 85.8 | 78.3 | 13.5 |
| 17 | A2-P-Sol-4 | A2-P-Sol-4-loops | 73.2 | 84.2 | 67.2 | 51.7 | 91.9 | 90.1 | 88.7 | 78.1 | 14.9 |
| 18 | A2-NP-Sol-2 | A2-P-Sol-4-loops | 74.7 | 77.1 | 64.6 | 55.8 | 95 | 89.9 | 87.9 | 77.9 | 14.2 |
| 19 | A2-NP-Sol-1 | A2-P-Sol-4-variables | 75.1 | 80.9 | 66.7 | 56.7 | 89.6 | 86.5 | 83 | 76.9 | 11.7 |
| 20 | A2-P-Sol-3-loops | A2-P-Sol-4-variables | 74.7 | 79.3 | 62.6 | 49.6 | 94.4 | 90.2 | 86.8 | 76.8 | 16 |
| 21 | A2-NP-Sol-2 | A2-P-Sol-3-loops | 73.7 | 76 | 59.9 | 49.6 | 93.4 | 88.4 | 85.4 | 75.2 | 15.8 |
| 22 | A2-NP-Sol-1 | A2-P-Sol-4-loops | 66.3 | 77.5 | 61.3 | 47.7 | 92.9 | 89.7 | 87.5 | 74.7 | 16.9 |

trained on the trainset, a prediction as 'plagiarized' or 'non-plagiarized' copies can be made on a pair of solutions.

The dataset for the task in hand is made as follows. Similarity scores between all solutions are computed (excluding the cases when solution *i* compared with itself). An instance in the supervised learning dataset is produced as follows:

- The attributes are formed by computing the similarity scores from the seven similarity measures for each assignment pair.
- The label (the correct answer) is one of the values: P (plagiarized) or NP (non-plagiarized).

We look up two solutions in hand and label is added manually to form the gold standard dataset. The dataset details are shown in Table III.

To build the framework for plagiarism detection, we used eight classification algorithms, namely, C4.5 decision tree, Logistic regression (two versions), Multilayer Perceptron, Naïve Bayes, OneR, Random Forest, and Support Vector Machines (SVM). We used the open-source WEKA workbench implementations of these algorithms [37]. Since the total number of instances is quite low, we used 10-fold cross-validation in all the experiments to evaluate the classification algorithms.

A prediction made by a classification algorithm can be compared with the corresponding reference instance from the dataset. We evaluate the predictions on three widely used measures: precision (P), recall (R), and F1, defined as below:

$$P = \frac{TP}{TP + FP} \qquad R = \frac{TP}{TP + FN} \qquad F_1 = \frac{2 \times P \times R}{P + R}$$

TP is True Positive, FP is False Positive, FN is False Negative, and F1 is the harmonic mean of P and R given equal weighting. The results are shown in Table I. It can be seen that Random Forest outperforms others, having an F1 score of 0.955, whereas, C4.5 is a close runner up.

*A. The Synthetic dataset*

The dataset formed from the IPCA corpus is imbalanced. It is mainly because when computed all possible pairs most of these are non-plagiarized. We make it balanced using Synthetic Minority Oversampling Technique (SMOTE) [38]. SMOTE over-samples the minority class by generating synthetic examples, which in return help a classifier to generalize better. We over-sample the minority class (i.e., Plagiarized instances) to 235% resulting in the dataset of 444 total instances, where non-plagiarized instances remain the same (223), and plagiarized instances increase to 221, as shown in Table III.

The classifiers are trained again on the oversampled dataset, and the results are reported in Table IV. It can be seen that Random Forest outperforms others, having an F1 score of 0.971, whereas C4.5 is a runner-up. The result on this oversampled dataset is increased by 0.016.

*B. Feature Importance*

A question worth asking is: which of the similarity methods play a decisive role in deciding if a set of assignment solutions is plagiarized or not. Since each method is represented as an attribute/feature in the supervised learning setting, we want to compute the importance of each feature. As shown in Table V, we employed Correlation-based Feature

Selection algorithm that returns each method's rank. It can be seen that Longest Common Subsequence and greedy string tiling with n=2 ranked highest in deciding if a pair of assignment solutions is plagiarized or not. On the original dataset, greedy string tiling with n=3 and Ngram overlap with n=2 also seems significant.

TABLE III. DATASETS DETAILS

|  | Non-Plagiarized | Plagiarized | Total |
|---|---|---|---|
| Original Dataset | 223 | 66 | 289 |
| Oversampled Dataset | 223 | 221 | 444 |

TABLE IV. A SUMMARY OF CLASSIFICATION RESULTS

|  | Original Dataset | | | Oversampled Dataset | | |
|---|---|---|---|---|---|---|
| Classifier | P | R | F1 | P | R | F1 |
| Random Forest | 0.955 | 0.955 | 0.955 | 0.971 | 0.971 | 0.971 |
| C4.5 | 0.948 | 0.948 | 0.948 | 0.953 | 0.953 | 0.953 |
| OneR | 0.92 | 0.92 | 0.917 | 0.912 | 0.912 | 0.912 |
| Multilayer Perceptron | 0.903 | 0.903 | 0.903 | 0.89 | 0.89 | 0.89 |
| Logistic | 0.898 | 0.9 | 0.899 | 0.896 | 0.894 | 0.894 |
| Simple Logistic | 0.891 | 0.893 | 0.892 | 0.903 | 0.901 | 0.901 |
| SVM | 0.888 | 0.889 | 0.889 | 0.911 | 0.908 | 0.908 |
| Naïve Bayes | 0.903 | 0.875 | 0.882 | 0.9 | 0.896 | 0.896 |

The results are from 0 to 1.

TABLE V. THE ROLE OF FEATURES

| Original Dataset | | Oversampled Dataset | |
|---|---|---|---|
| LCS | 10 | LCS | 10 |
| GST2 | 10 | GST2 | 10 |
| N2 | 9 | N1 | 8 |
| GST3 | 9 | GST3 | 7 |
| N1 | 4 | N2 | 7 |
| GST1 | 2 | N3 | 7 |
| N3 | 1 | GST1 | 6 |

Results between 0 to 10 (0: least significant, 10: most significant). LCS: Longest Common Subsequence; N1, N2, N3: N-gram overlap n=1, n=2, n=3 respectively; GST1, GST2, GST3: Greedy String Tiling with n=1, n=2, n=3 respectively

## VI. CONCLUSION

In this paper, we proposed a supervised learning framework for the detection of code reuse in the assignments of the introductory programming course. Our results demonstrate that supervised learning using token based similarity measures as attributes has great potential for plagiarism detection. With a small hand-built corpus of programming assignments from the introductory course named IPCA, we get excellent F1 scores. The results improve further when a synthetic dataset is used. We found that Longest Common Subsequence and Greedy String Tiling with n=2 are the best features for the task. Future work includes extending the framework to other programming languages such as Python, Java, and C#. Another venue of future work is to support advanced programming courses, which require a large programming assignments corpus covering advanced topics.